\def\year{2022}\relax
\documentclass[letterpaper]{article} 
\usepackage{aaai22}  
\usepackage{times}  
\usepackage{helvet}  
\usepackage{courier}  
\usepackage[hyphens]{url}  
\usepackage{graphicx} 
\usepackage{natbib}  
\usepackage{caption} 
\DeclareCaptionStyle{ruled}{labelfont=normalfont,labelsep=colon,strut=off} 
\usepackage{color}
\usepackage{algorithm}
\usepackage{amsmath}
\usepackage{amssymb}
\usepackage{algorithmic}
\usepackage[switch]{lineno}
\definecolor{grey}{rgb}{0.5, 0.5, 0.5}
\usepackage{hyperref}

\usepackage{newfloat}
\usepackage{listings}
\floatstyle{ruled}
\newfloat{listing}{tb}{lst}{}
\floatname{listing}{Listing}
\title{Uncertainty Estimation via Response Scaling for Pseudo-mask Noise Mitigation in Weakly-supervised Semantic Segmentation}

\author {
	Yi Li\textsuperscript{\rm 1,2},
	Yiqun Duan \textsuperscript{\rm 3},
    Zhanghui Kuang \textsuperscript{\rm 2},
    Yimin Chen \textsuperscript{\rm 2},
    Wayne Zhang \textsuperscript{\rm 2},
    Xiaomeng Li\textsuperscript{\rm 1,4,}\thanks{Corresponding Author: {\tt\small eexmli@ust.hk}}
}
\affiliations {
	\textsuperscript{\rm 1} Department of Electronic and Computer Engineering, The Hong Kong University of Science and Technology \\
	\textsuperscript{\rm 2} SenseTime Research 
	\textsuperscript{\rm 3} Department of Computer Science, University of Technology Sydney\\
	\textsuperscript{\rm 4} The Hong Kong University of Science and Technology Shenzhen Research Institute
}

\usepackage{bibentry}

\begin{document}

\maketitle
\newcommand{\xmli}[1]{{\color{blue}{[XM: #1]}}}

\begin{abstract}

Weakly-Supervised Semantic Segmentation (WSSS) segments objects without a heavy burden of dense annotation. While as a price, generated pseudo-masks exist obvious noisy pixels, which result in sub-optimal segmentation models trained over these pseudo-masks. But rare studies notice or work on this problem, even these noisy pixels are inevitable after their improvements on pseudo-mask. So we try to improve WSSS in the aspect of noise mitigation. And we observe that many noisy pixels are of high confidence, especially when the response range is too wide or narrow, presenting an uncertain status. Thus, in this paper, we simulate noisy variations of response by scaling the prediction map multiple times for uncertainty estimation. The uncertainty is then used to weight the segmentation loss to mitigate noisy supervision signals. We call this method URN, abbreviated from Uncertainty estimation via Response scaling for Noise mitigation. Experiments validate the benefits of URN, and our method achieves state-of-the-art results at 71.2\% and 41.5\% on PASCAL VOC 2012 and MS COCO 2014 respectively, without extra models like saliency detection. Code is available at \href{https://github.com/XMed-Lab/URN}{https://github.com/XMed-Lab/URN}.

\end{abstract}

\section{Introduction}


\noindent Semantic segmentation is the fundamental task in computer vision. 
One of the main challenges is the prohibitive cost of obtaining dense pixel-level annotations to supervise model training. 
To reduce the annotation cost, researchers replace the dense pixel-level annotations by weak annotations, such as scribble~\cite{lin2016scribblesup,Tang_2018_CVPR}, bounding boxes \cite{Xu_2015_CVPR,dai2015boxsup,Khoreva_2017_CVPR}, points \cite{bearman2016s} and image-level labels~\cite{kolesnikov2016seed,pathak2015constrained,pinheiro2015image,ahn2018learning}. 
In this paper, our goal is to develop a novel method for Weakly-Supervised Semantic Segmentation (WSSS) based on image-level class labels only. 


Most existing methods~\cite{Wang_2020_CVPR,zhang2020causal,li2021pseudo} in WSSS follow the common pipeline that generates pseudo-masks firstly and then trains the segmentation model with the pseudo-masks in a fully-supervised manner. 
These methods improve the quality of pseudo-mask by solving the under activation problem of Class Activation Map \cite{zhou2016learning} as these papers \cite{kolesnikov2016seed,jiang2019integral,chang2020weakly}, also distinguishing foreground from background with AffinityNet \cite{ahn2018learning} or saliency \cite{yao2020saliency,fan2020learning}. But the noisy pixels are inevitable after applying these methods. As far as we know, only PMM \cite{li2021pseudo} tries to relieve the influence of noise. While it does not work on refined pseudo-mask and COCO dataset, without location of noisy pixels. So in this paper, we mitigate the noise in a more effective way.

\begin{figure}[t]
\begin{centering}
\includegraphics[width=2.7cm,height=2.3cm]{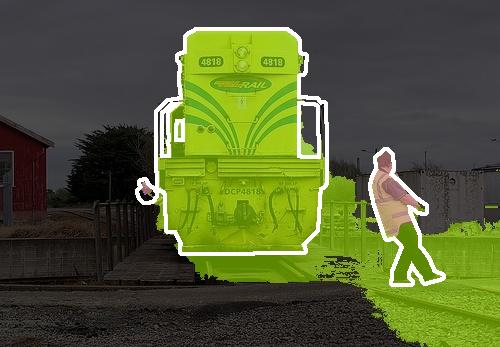} \includegraphics[width=2.7cm,height=2.3cm]{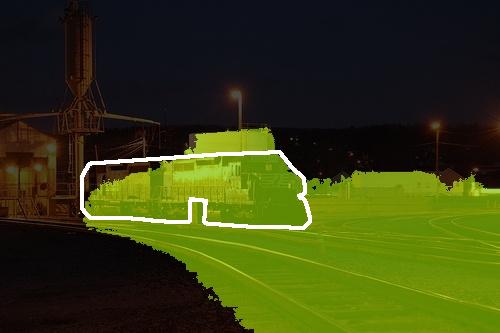} \includegraphics[width=2.7cm,height=2.3cm]{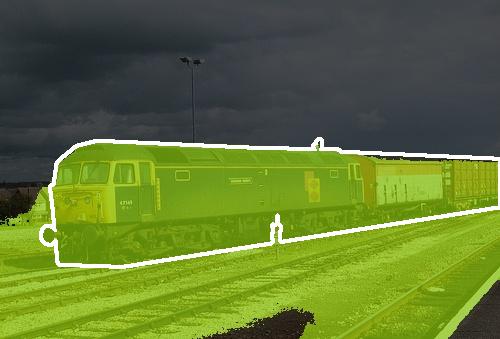}
\end{centering}
\begin{centering}
\includegraphics[width=2.7cm,height=2.3cm]{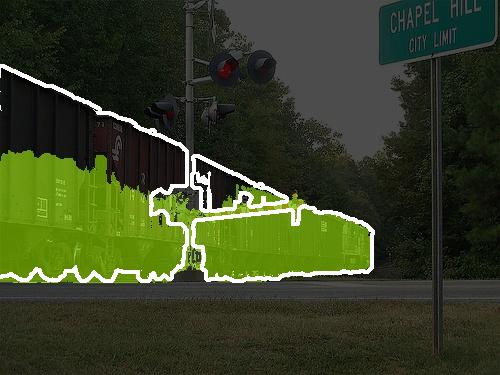} \includegraphics[width=2.7cm,height=2.3cm]{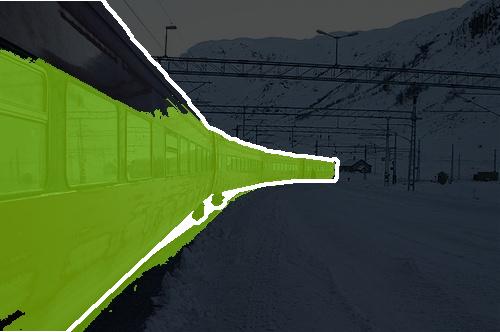} \includegraphics[width=2.7cm,height=2.3cm]{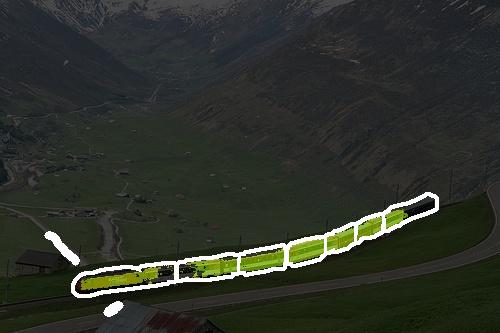}
\end{centering}
\begin{centering}
\caption{False pixels (noise) in pseudo-masks of the state-of-the-art WSSS method \cite{li2021pseudo}. Top: wide response scales cause false positives. Bottom: narrow response scales make missing positives. The white line is boundary of ground-truth.}
\label{fig:range} 
\par\end{centering}
\centering{} 
\end{figure}

We observe that the noisy pixels are mainly divided into missing positives and false positives. To be specific, if the response range (scope of activated area) is narrow, the missing positives take the lead, otherwise false positives become the majority as shown in Fig.\ref{fig:range}. Because of the variation of the response scale, there are many false pixels (noise) with high confidence. And this phenomenon is exactly belongs to uncertainty. Thus, in this paper, we locate the noise via uncertainty estimation.


\begin{figure}[t]
\begin{centering}
\includegraphics[width=2.7cm,height=2.1cm]{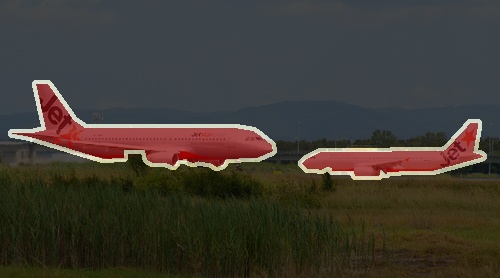} \includegraphics[width=2.7cm,height=2.1cm]{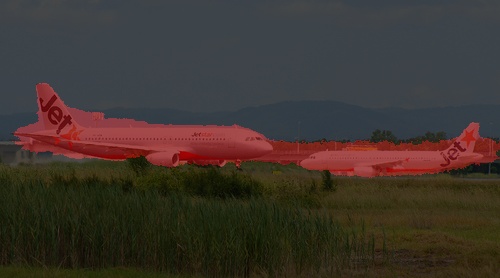} \includegraphics[width=2.7cm,height=2.1cm]{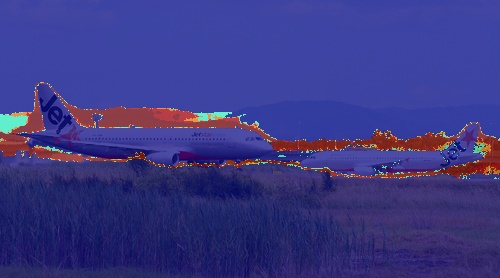} \end{centering}
\begin{centering}
\includegraphics[width=2.7cm,height=2.1cm]{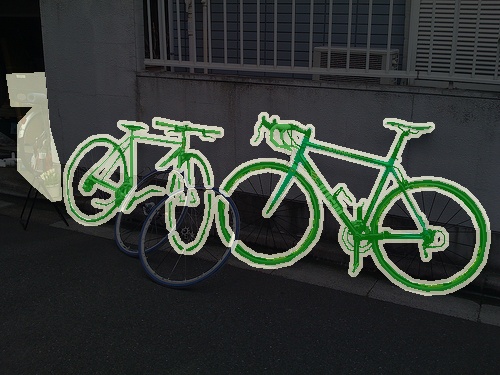} \includegraphics[width=2.7cm,height=2.1cm]{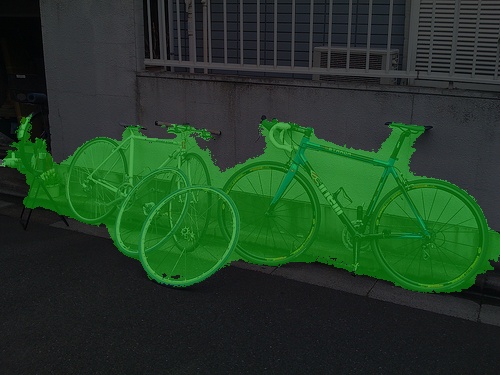} \includegraphics[width=2.7cm,height=2.1cm]{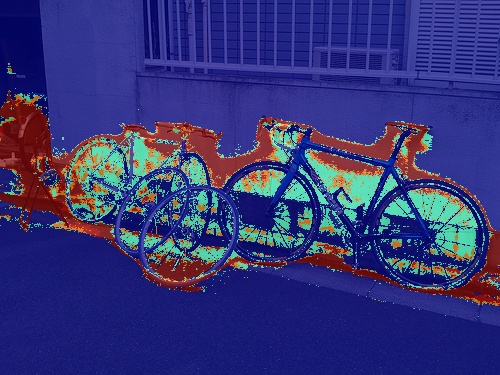} \end{centering}
\begin{centering}
\includegraphics[width=2.7cm,height=2.1cm]{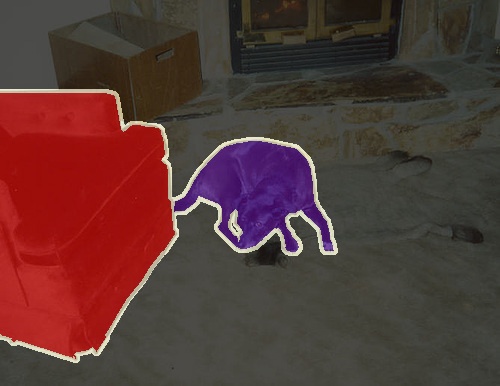} \includegraphics[width=2.7cm,height=2.1cm]{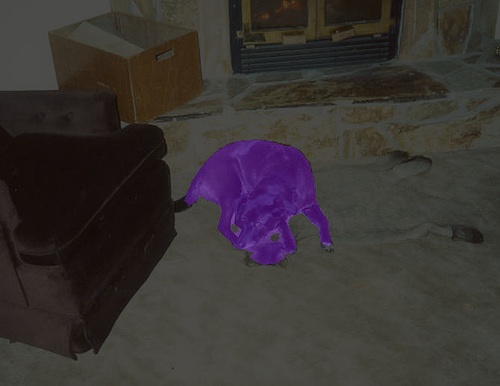} \includegraphics[width=2.7cm,height=2.1cm]{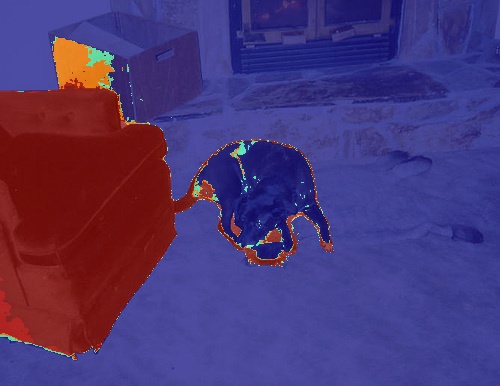} \end{centering}
\begin{centering}
\caption{Uncertainty estimated by response scaling is positively correlated with false pixels (noise) of pseudo-mask. Left: ground truth. Middle: noisy pseudo-mask. Right: uncertain map after normalization.}
\label{uncertainty} 
\par\end{centering}
\centering{} 
\end{figure}

Uncertainty estimation requires multiple predictions for each image, so how to generate these predictions and reflect the noise is important in the setting of WSSS. By observation of false pixels whose confidences are high in Figure~\ref{fig:range}, the uncertain pixels are highly related to the response scale. So we simulate response maps in different activation scales by probability scaling. Besides, we add dense-CRF~\cite{zheng2015conditional} to response maps after scaling scheme as the formation of pseudo-mask. After above operations, we estimate the uncertainty with these scaled responses, and the affinity is built between uncertainty and noise. 

After uncertainty estimation via response scaling, we normalize the uncertainty and transform it to loss weight in optimization to mitigate the damage of noise. We observe that the pixel whose uncertainty is higher is more likely to be noise as shown in Figure~\ref{uncertainty}. Thus, a lower loss weight is applied to relieve the adverse effect of it. We multiply the weight map and loss map with a mean operation to get the final loss in optimization.

The overall process is call URN abbreviated from Uncertainty estimation via Response scaling for Noise mitigation. URN is practical and effective for the segmentation with noisy pseudo-mask from weakly supervision. We validate the proposed URN with baseline method PMM~\cite{li2021pseudo} on a prevailing dataset PASCAL VOC 2012~\cite{everingham2010pascal} and a challenging dataset MS COCO 2014 with image-level supervision. And we get new state-of-the-art results at 71.2\% on VOC and 41.5\% on COCO in metric mIoU.

We summarize the contributions as follows :
\begin{itemize}
\item We observe that false pixels (noise) with high confidences are related to the scale of response range, and we locate these uncertain pixels via response scaling.
\item We originally mitigate the damage of noise in optimization of Weakly-Supervised Semantic Segmentation via loss weight from uncertainty. 
\item We achieve new state-of-the-art results on main datasets in Weakly-Supervised Semantic Segmentation.
\end{itemize}

\section{Related Work}
\textbf{Weakly-Supervised Semantic Segmentation.} 
In recent years, considerable efforts have been devoted to developing label-efficient semantic segmentation methods~\cite{lin2016scribblesup,Tang_2018_CVPR,Xu_2015_CVPR,dai2015boxsup,Khoreva_2017_CVPR,kolesnikov2016seed,pathak2015constrained,pinheiro2015image,ahn2018learning,Wang_2020_CVPR,yu2019uncertainty,li2020transformation,li2018semi}.
Among these methods, image-level class labels require the least annotation cost. 
Existing WSSS methods mainly adopt two-stage procedures. The first stage optimizes class activation maps (CAMs)~\cite{zhou2016learning} produced by a multi-label classification network to generate pseudo ground-truth.  
The second stage is to train a fully supervised semantic segmentation network via pseudo ground-truth. 
To improve WSSS, most algorithms focus on improving the quality of CAMs through expanding partial response area to whole foreground region~\cite{kolesnikov2016seed,jiang2019integral,chang2020weakly}.
Some methods distinguish foreground from background to improve the quality of CAMs by self-supervised method~\cite{ahn2018learning,8953768} or introducing extra information liking saliency detection model \cite{lee2019ficklenet,yao2020saliency,jiang2019integral,yao2020saliency,li2020group,fan2020learning} and object proposals~\cite{liu2020leveraging}.

In WSSS, researchers mainly focus on generating high-quality pseudo labels to improve the performance, less attention has been paid on \emph{how to improve the segmentation model with these imperfect labels.} In this paper, We realize that since noisy pixels are inevitable, the appropriate way is to mitigate the damage of uncertain pixels.



\noindent \textbf{Uncertainty in Segmentation.}
Deep neural networks provide a probability with each prediction, occasionally, it is false even the probability is high. In paper \cite{kendall2017uncertainties}, this phenomenon is called epistemic uncertainty resulting from the model itself. 
Early works use Monte-Carlo Dropout \cite{kendall2015bayesian,gal2016dropout,pmlr-v37-blundell15,pmlr-v48-gal16} to approximate the posterior distribution for uncertainty estimation, also, Ensemble~\cite{lakshminarayanan2017} and multi-head~\cite{lee2015m, lee2016stochastic, rupprecht2017learning} are also applied. 
Follow-up researchers majorly concentrate on proposing improved approaches to acquire better output uncertainty maps.  
Here, inference-based~\cite{kendall2017uncertainties, tanno2017bayesian, jungo2020analyzing} mostly utilize Bayesian Inference or Stochastic Inference to improve the quality of uncertainty map, meanwhile auto-encoder-based~\cite{kingma2013auto, sohn2015} methods introduce reconstruction to acquire better spatial precision of the uncertainty map. 
These papers mostly predict uncertainty masks to facilitate human's judgement, such as it in medical image segmentation scenarios. However, previous methods mostly emphasize on improving the quality of the uncertainty map but rarely explore how to improve the segmentation precision through uncertainty. Instead, they assemble the predictions for better results.

\noindent Different from previous works that estimate uncertainty for visualization or annotation, we utilize uncertainty to mitigate noise in optimization of segmentation for better performance. And the uncertainty is estimated based on the observation of noise's characteristics in Weakly-supervised Semantic Segmentation.




\noindent \textbf{Learning with Noisy Labels.}
Our focus in this paper is to mitigate noisy labels in WSSS. Here, we discuss some related work in learning with noisy labels.
Existing methods of learning with noisy labels could be classified into four categories. 1) Label correction methods propose to improve the quality of the labels by applying label corrections via using complex noise prediction models such as graphical models~\cite{xiao2015learning}, conditional random fields~\cite{vahdat2017toward}, and neural networks \cite{lee2017cleannet,veit2017learning}. 
2) Label-weighted loss methods modify loss functions during training, based on label-dependent weights \cite{natarajan2013learning,goldberger2016training,sukhbaatar2014training} or estimated noise transition matrix that defines the probability of mislabeling one class with another~\cite{han2018masking, patrini2017making,reed2014training, pereyra2017regularizing}. For example, \citet{xu2019l_dmi} introduces a Mutual Information (MI) loss for robust fine-tuning of a CE pre-trained model. 3) Fine-grained label correction in training designs adaptive training strategies that are more robust to noisy labels. 
These methods mostly correct labels by teacher-student distillation~\cite{jiang2018mentornet,yu2019does,kumar2019secost} and apply jointly training or learning-mutually mechanism between parallel models~\cite{han2018co, tanaka2018joint, kim2019nlnl}. 4) Robust loss functions also are proposed as more generic solution for robust learning by proposing and proving new loss function, such as Mean Absolute Error (MAE) losses~\cite{ghosh2017robust}, gradient saturation MAE~\cite{zhang2018generalized}, Generalized Cross-Entropy (GCE) loss \cite{zhang2018generalized}, and Symmetric Cross-Entropy (SCE) \citet{wang2019symmetric}. Empirically justified approaches that directly modify the magnitude of the loss gradients are also an active line of research~\cite{wang2019imae, wang2019derivative}.

\noindent Unlike the previous reweight methods based on probability, there are many pixels with high confidences in pseudo-masks, presenting an uncertain status. Thus, for the first time in WSSS, we propose a new method to estimate the uncertainty to locate the possible noise for mitigation. Besides, we present a distillation scheme based on pseudo-masks in WSSS.

\section{Methodology}
\begin{figure*}[ht]
\begin{centering}
\includegraphics[scale=0.39]{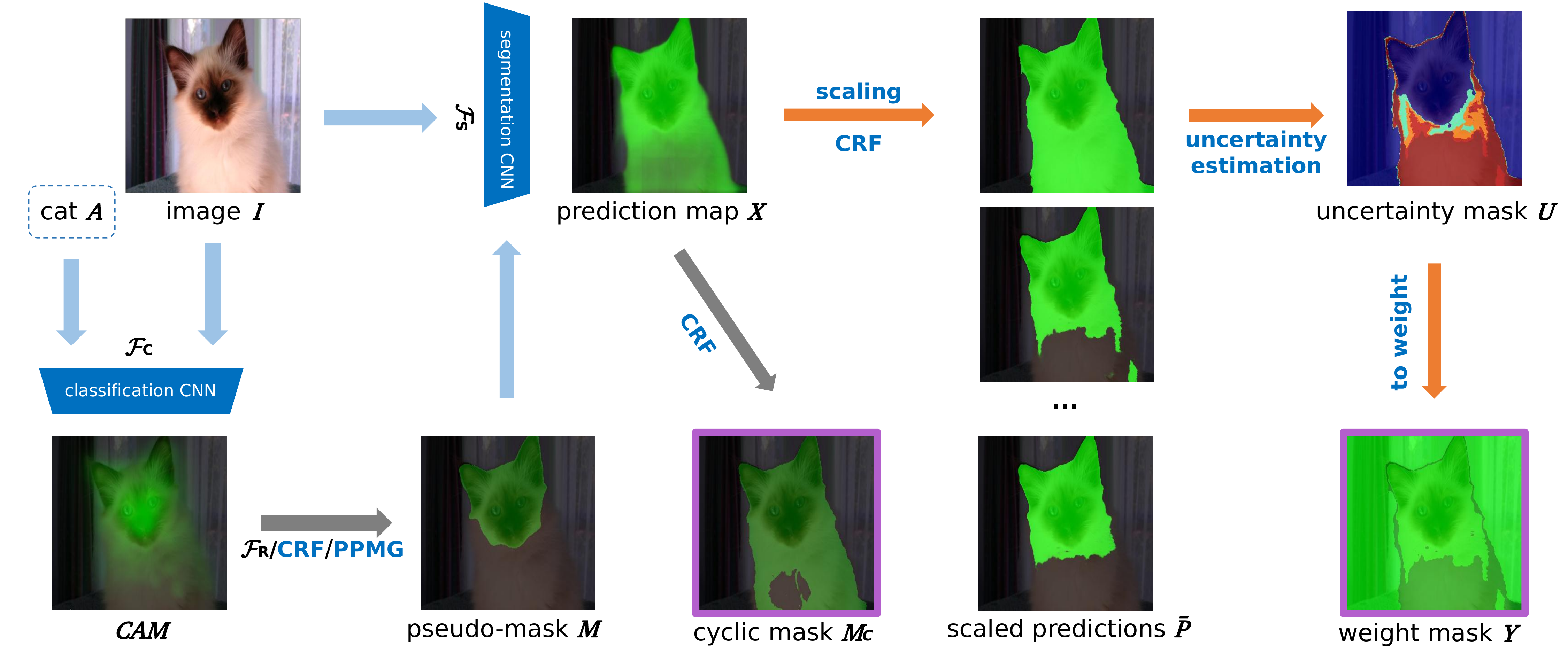}
\par\end{centering}
\caption{\label{fig:process}Process of URN. Outputs are weight mask $\boldsymbol{Y}$ and cyclic mask $\boldsymbol{M}_{c}$ with purple borders. Blue arrays indicate the training inputs of CNNs, and orange arrays belong to the operations of URN. $\mathcal{F}_{R}$, CRF and PPMG are optional post-processes. The weight mask looks like the intermediate between pseudo-mask and cyclic mask in this image.}
\end{figure*}

In this section, we firstly introduce the pipeline of Weakly-supervised Semantic Segmentation and where our method is applied. Then we elaboration the proposed Uncertainty Estimation via Response Scaling. In the last, we describe the transformation from uncertainty to loss weight. We depict the whole process as Fig.\ref{fig:process} for better understanding.

Let's define the segmentation function with image $\boldsymbol{I\in\mathbb{R}^{H\times W}}$ as $\mathcal{F}_{S}(\boldsymbol{I})$ which trained from pseudo-mask $\boldsymbol{M\in\mathbb{R}^{H\times W}}$, and its prediction $\boldsymbol{X}\in\mathbb{R}^{C\times H\times W}$ is the input of the proposed method Uncertainty estimation via Response scaling for Noise mitigation (URN) $Urn(\boldsymbol{X})$ whose output is loss weight $\boldsymbol{Y}\in\mathbb{R}^{C\times H\times W}$ for noise mitigation in segmentation optimization.

\subsection{URN in Weakly-supervised Semantic Segmentation}
Latest Weakly-supervised Semantic Segmentation has three main components, namely classification, pseudo-mask refinement and segmentation. We name them as $\mathcal{F}_{C}(\boldsymbol{I}, \boldsymbol{L})$, $\mathcal{F}_{R}(\boldsymbol{M}, \boldsymbol{I})$, $\mathcal{F}_{S}(\boldsymbol{I}, \boldsymbol{M})$ respectively. Note that the first variable is input of inference and the second variable is ground-truth in training. $\mathcal{F}_{C}$ is trained with image-level annotation $\boldsymbol{A}$ and converts image $\boldsymbol{I}$ to Class Activation Map $\boldsymbol{CAM}\in\mathbb{R}^{C\times H\times W}$ via operation $Cam$ as Eq.\ref{eq:to_cam}
\begin{equation}
\boldsymbol{CAM} = Cam(\mathcal{F}_{C}(\boldsymbol{I}))
\label{eq:to_cam}
\end{equation}

Then $\boldsymbol{CAM}$ is refined by $\mathcal{F}_{R}$ to obtain pseudo mask $\boldsymbol{M}$. In most methods $\mathcal{F}_{R}$ is dense-CRF and in our baseline it is PPMG \cite{li2021pseudo}. In paper \cite{ahn2018learning} it is AffinityNet with Random Walk to learn the foreground and background. We collectively referred them to $\mathcal{F}_{R}$ as:
\begin{equation}
\boldsymbol{M} = \mathcal{F}_{R}(\boldsymbol{CAM})
\end{equation}

In this paper, we focus on the last stage $\mathcal{F}_{S}$ which is trained from $\boldsymbol{M}$ and inference with image. Our improvement is an additive loss weight $\boldsymbol{Y}$ from Uncertainty Estimation via Response Scaling with a transformation operation. We conclude these operations as $Urn$:
\begin{equation}
\boldsymbol{Y} = Urn(\mathcal{F}_{S}(\boldsymbol{I}))
\end{equation}

After $Urn$ calculating the weight mask, $\bar{\mathcal{F}_{S}}$ is trained with Cyclic Pseudo-mask $\boldsymbol{M}_{c}$ in PMM \cite{li2021pseudo}, image $\boldsymbol{I}$ and weight mask $\boldsymbol{Y}$ for the final deployment.

\subsection{Uncertainty Estimation via Response Scaling}
Bayesian neural networks \cite{denker1990transforming,mackay1992practica} set deep model into a probabilistic
model by learning the distribution of prediction. While Bayesian neural network’s prediction is hard to obtain. Thus, they apply varied inference to approximate the posterior of the model via dropout, such as Monte Carlo Dropout \cite{kendall2017uncertainties} or ensemble.


In WSSS, our goal is to measure the variance of pseudo-masks due to the inevitable noise on pseudo-masks. But predictions from dropout or different models cannot directly obtain pseudo-masks in the behavior of noise. Activation ranges are likely too narrow or too wide simultaneously, because the operation target is the network weight instead of the response itself. We observe that the noise in pseudo-mask of segmentation is always related to the response range of foregrounds. Motivated by this observation, we propose the Uncertainty Estimation via Response Scaling to mitigate the noise. Scaling operation with CRF could simulate the true behavior of noisy pseudo-masks. There, the variance of this noisy pseudo-masks in real distribution is the uncertainty that reflect the variance of real noise in pseudo-masks.

We start the estimation from segmentation model $\mathcal{F}_{S}$ which is trained from pseudo-masks, we have its prediction feature map as:
\begin{equation}
\boldsymbol{X} = \mathcal{F}_{S}(\boldsymbol{I})
\label{eq:forward}
\end{equation}

We firstly record the existed labels in pseudo-mask $\boldsymbol{M}$, and these labels are corresponded to the indexes $\mathbb{R}^{\bar{C}}$ of $\boldsymbol{X}$ in channel dimension. We only keep the appeared channels to speed up the process. we have the selected feature map $\boldsymbol{\bar{X}}$ as:
\begin{equation}
\boldsymbol{\bar{X}} = \boldsymbol{X}_{(\mathbb{R}^{\bar{C}},:,:)},\mathrm{ s.t. },\mathbb{R}^{\bar{C}}\in\boldsymbol{M}
\label{eq:start}
\end{equation}

Then we apply the scaling scheme to $\boldsymbol{\bar{X}}$ via exponential function with varied power factors $S\in\mathbb{R}^{N}$ on the target channel to generate varied response maps, and in this process, other channels keep the same. When the scale factor is upper than 1, the response scale of target class declines and the number of false positive reduces too. If it's lower than 1 the scale expands and there are less missing positive pixels. For each label in $\mathbb{R}^{\bar{C}}$, we adjust the scale of one channel and keep others the same for category specifically variation. We have scaled predictions $\boldsymbol{P}\in\mathbb{R}^{\bar{C}\times N\times \bar{C}\times H\times W}$ for each class and scale as:
\begin{equation}
\boldsymbol{P}_{(\bar{c},n,\bar{c},:,:)} = \boldsymbol{\bar{X}}_{(\bar{c},:,:)}^{S_{n}}, \forall{n,\bar{c}}
\end{equation}

As we want to estimate the uncertainty of pseudo-mask predicted from segmentation, we introduce dense-CRF function $Crf$ and argmax $Arg$ to generate pseudo-masks in varied scales as follows:
\begin{equation}
\boldsymbol{\bar{P}}_{(\bar{c},n,:,:,:)} = Crf(\boldsymbol{\bar{P}}_{(\bar{c},n,:,:,:)}, \boldsymbol{I}), \forall{n,\bar{c}}
\label{eq:scaling}
\end{equation}

$\boldsymbol{\bar{P}}$ indicates the scaled predictions after dense-CRF, then we get the pseudo-masks in varied scales and classes $\boldsymbol{\bar{M}}\in\mathbb{R}^{\bar{C}\times N\times H\times W}$ by reducing the third category channel (ranking from 0) via argmax:
\begin{equation}
\boldsymbol{\bar{M}} = Arg(\boldsymbol{\bar{P}}, dim=2)
\end{equation}

After get scales pseudo-masks we start to estimate the uncertainty, we select variance $\sigma^2\in\mathbb{R}^{\bar{C}\times H\times W}$ as the metric to approximate the uncertainty for its simplicity and versatility. We calculate $Var$ in second dimension (scaling factors) $N$ for each foreground categories:
\begin{equation}
Var(\boldsymbol{\bar{M}}) = \frac{1}{N-1}\sum_{n=0}^{N} (\boldsymbol{\bar{M}}_{(\bar{c},n,:,:)} - \mathbb{E}(\boldsymbol{\bar{M}}_{\bar{c},:,:,:}))^2, \forall{\bar{c}}
\end{equation}

Finally, we apply $Max$ operation in category dimension as Eq.\ref{eq:max} to get class agnostic uncertainty map $\boldsymbol{U}\in\mathbb{R}^{H\times W}$ for the later loss reweight with min-max normalization $Norm$.
\begin{equation}
\boldsymbol{U} = Norm(Max(Var(\boldsymbol{\bar{M}}), dim=0))
\label{eq:max}
\end{equation}

\subsection{Noise Mitigation in Optimization}

Reweight scheme is widely usage in computer vision tasks like classification and detection to balance category or adjusting the importance of certain kind of samples. Following the mechanism, we deploy a lower loss weight for the possible noise pixels based on the uncertainty estimated by response scaling to mitigate the damage of noise in training.

Since pixels with high uncertainty deserve low loss weight to limit its influence. We set initial weight mask $\boldsymbol{W}\in\mathbb{R}^{H\times W}$ as:
\begin{equation}
\boldsymbol{W} = 1 - \boldsymbol{U}
\end{equation}

To make the weight mask more controllable, we assign certain pixels and uncertain pixels via threshold $t$ guided by $\boldsymbol{W}$. And we mitigate the impact of possible noise by a lower loss weight. We define the final weight mask $\boldsymbol{Y}$ as:
\begin{equation}
\boldsymbol{Y}_{(h,w)}=\begin{cases}
1 & \boldsymbol{W}_{(h,w)}>=t\\
t & \boldsymbol{W}_{(h,w)}<t
\end{cases}\forall{h,w}
\label{eq:end}
\end{equation}

In the end, we multiply the weight mask $\boldsymbol{Y}$ to the loss mask $\boldsymbol{L}\in\mathbb{R}^{H\times W}$ from cross entropy loss with target cyclic pseudo-mask $\boldsymbol{M}_{c}$ and prediction map $\boldsymbol{X}$. We have the segmentation loss in as Eq.(\ref{eq:loss}):
\begin{equation}
\boldsymbol{L}(\boldsymbol{X}, \boldsymbol{M}_{c})_{(h,w)} = -\boldsymbol{Y}_{(h,w)} \cdot log\frac{exp(\boldsymbol{X}_{({\boldsymbol{M}_{c}}_{(h,w)},h,w)})}{\sum_{c=0}^{C}exp(\boldsymbol{X}_{c,h,w})},\forall{h,w}
\label{eq:loss}
\end{equation}
\begin{table*}[h]
\begin{centering}
\begin{tabular}{ccc|ccc}
\hline 
Method & Backbone & Supervision & VOC12 val & VOC12 test & COCO14 val\tabularnewline
\hline 
\hline 
BFBP \cite{saleh2016built} & \textcolor{grey}{VGG16} & $\mathcal{I}$ & \textcolor{grey}{46.6$^{\dag}$} & \textcolor{grey}{48.0$^{\dag}$} & \textcolor{grey}{20.4$^{\dag}$}\tabularnewline
    SEC \cite{kolesnikov2016seed} & \textcolor{grey}{VGG16} & $\mathcal{I}$ & \textcolor{grey}{50.7$^{\dag}$} & \textcolor{grey}{51.7$^{\dag}$} & \textcolor{grey}{22.4$^{\dag}$}\tabularnewline
AffinityNet \cite{ahn2018learning} & \textcolor{grey}{ResNet-38} & $\mathcal{I}$ & \textcolor{grey}{61.7} & \textcolor{grey}{63.7} & -\tabularnewline
IRNet \cite{8953768} & \textcolor{grey}{ResNet-50} & $\mathcal{I}$ & \textcolor{grey}{63.5} & \textcolor{grey}{64.8} & -\tabularnewline
OAA \cite{jiang2019integral} & ResNet-101 & $\mathcal{I}$ & 63.9 & 65.6 & -\tabularnewline
ICD \cite{fan2020learning} & ResNet-101 & $\mathcal{I}$ & 64.1 & 64.3 & -\tabularnewline
{SEAM \cite{Wang_2020_CVPR}} & \textcolor{grey}{ResNet-38} & $\mathcal{I}$ & \textcolor{grey}{64.5} & \textcolor{grey}{65.7} & \textcolor{grey}{31.7} -\tabularnewline
SSDD \cite{shimoda2019self} & ResNet-101 & $\mathcal{I}$ & 64.9 & 65.5 & -\tabularnewline
CONTA \cite{zhang2020causal} & \textcolor{grey}{ResNet-38} & $\mathcal{I}$ & \textcolor{grey}{66.1} & \textcolor{grey}{66.7} & \textcolor{grey}{32.8}\tabularnewline
SC-CAM \cite{chang2020weakly} & ResNet-101 & $\mathcal{I}$ & 66.1 & 65.9 & -\tabularnewline
Sun et al. \cite{sun2020mining} & ResNet-101 & $\mathcal{I}$ & 66.2 & 66.9 & -\tabularnewline
PMM\cite{li2021pseudo} & \textcolor{grey}{ResNet-38} & $\mathcal{I}$ & \textcolor{grey}{\underline{68.5}} & \textcolor{grey}{\underline{69.0}} & \textcolor{grey}{\underline{36.7}}\tabularnewline
PMM\cite{li2021pseudo} & \textcolor{grey}{ScaleNet-101} & $\mathcal{I}$ & \textcolor{grey}{\underline{67.1}} & \textcolor{grey}{\underline{67.7}} & \textcolor{grey}{\underline{40.2}}\tabularnewline
PMM\cite{li2021pseudo} & \textcolor{grey}{Res2Net-101} & $\mathcal{I}$ & \textcolor{grey}{\underline{70.0}} & \textcolor{grey}{\underline{70.5}} & \textcolor{grey}{\underline{35.7}}\tabularnewline
\hline
DSRG \cite{huang2018weakly} & ResNet-101 & $\mathcal{I}$+ESnE+MSRA-B & 61.4 & 63.2 & \textcolor{grey}{26.0$^{\dag}$}\tabularnewline
FickleNet \cite{lee2019ficklenet} & ResNet-101 & $\mathcal{I}$+$\mathcal{S}$ & 64.9 & 65.3 & -\tabularnewline
SDI \cite{Khoreva_2017_CVPR} & ResNet-101 & $\mathcal{I}$+$\mathcal{D}$+BSDS & 65.7 & 67.5 & -\tabularnewline
OAA$^{,}$ \cite{jiang2019integral} & ResNet-101 & $\mathcal{I}$+$\mathcal{S}$ & 65.2 & 66.4 & \tabularnewline
SGAN \cite{yao2020saliency} & ResNet-101 & $\mathcal{I}$+$\mathcal{S}$ & 67.1 & 67.2 & \underline{33.6}\tabularnewline
ICD \cite{fan2020learning} & ResNet-101 & $\mathcal{I}$+$\mathcal{S}$ & 67.8 & 68.0 & -\tabularnewline
Li et al. \cite{li2020group} & ResNet-101 & $\mathcal{I}$+$\mathcal{S}$ & \underline{68.2} & \underline{68.5} & \textcolor{grey}{28.4$^{\dag}$}\tabularnewline
LIID \cite{liu2020leveraging} & ResNet-101 & $\mathcal{I}$+SOP & 66.5 & 67.5 & -\tabularnewline
LIID \cite{liu2020leveraging} & \textcolor{grey}{Res2Net-101} & $\mathcal{I}$+SOP & \textcolor{grey}{69.4} & \textcolor{grey}{70.4} & -\tabularnewline
\hline
URN & \textcolor{grey}{ResNet-38} & $\mathcal{I}$ & \textcolor{grey}{\textbf{69.4}}$_{\textcolor{blue}{0.9}}$ & \textcolor{grey}{\textbf{70.6}}$_{\textcolor{blue}{1.6}}$ & \textcolor{grey}{\textbf{40.5}}$_{\textcolor{blue}{3.8}}$\tabularnewline
URN & ResNet-101 & $\mathcal{I}$ & \textbf{69.5}$_{\textcolor{blue}{1.3}}$ & \textbf{69.7}$_{\textcolor{blue}{1.2}}$ & \textbf{40.7}$_{\textcolor{blue}{7.1}}$\tabularnewline
URN & \textcolor{grey}{ScaleNet-101} & $\mathcal{I}$ & \textcolor{grey}{\textbf{70.1}}$_{\textcolor{blue}{3.0}}$ & \textcolor{grey}{\textbf{70.8}}$_{\textcolor{blue}{3.1}}$ & \textcolor{grey}{\textbf{40.8}}$_{\textcolor{blue}{0.6}}$\tabularnewline
URN & \textcolor{grey}{Res2Net-101} & $\mathcal{I}$ & \textcolor{grey}{\textbf{71.2}}$_{\textcolor{blue}{1.2}}$ & \textcolor{grey}{\textbf{71.5}}$_{\textcolor{blue}{1.0}}$ & \textcolor{grey}{\textbf{41.5}}$_{\textcolor{blue}{5.8}}$\tabularnewline
\hline 
\end{tabular}
\par\end{centering}
\caption{\label{tab:results}Performance companion with state-of-the-art WSSS methods on VOC 2012 and COCO 2014. The middle part lists the methods with extra supervision. $\mathcal{I}$, $\mathcal{S}$, $\mathcal{D}$ indicate supervisions of image-level tag, saliency, detection respectively. SOP is segment-based object proposals. Other extra information is about data. $^{\dag}$ indicates backbone of VGG. Results of ResNet-101 are in the color of black and others are of grey. For each backbone, the second result has an underline and the best is bold with gain in blue.}
\end{table*}

\section{Experiments}
\subsection{Datasets}
\textbf{PASCAL VOC 2012}: It is the most prevalent dataset in Weakly-supervised Semantic Segmentation, because of the moderate difficulty and quantity. To be specific, it is consisted of 20 foreground categories and one background class, divided into train set, validation set and test set at the quantities of 1464 images, 1449 images, 1456 images respectively. Besides the original data, additional images and annotations from SBD \cite{hariharan2011semantic} are used which is called trainaug set at number 10582. In WSSS, all the pixel-level annotations are converted to image-level multi-label annotations in classification phase.

\noindent \textbf{MS COCO 2012}: MS COCO is the main dataset in object detection, also some works in WSSS report their results on the version of 2012. It is a challenging dataset which provides more categories than PASCAL VOC, and it is smaller in average object size. MS COCO 14 dataset ranges from 0 to 90, among them, 80 categories are valid foreground with one background, and other 10 categories are not evaluated. The train set contains 82081 images and the number of validation set is 40137. Save to PASCAL VOC 2012, the evaluation metric is Mean intersection over union (mIoU).

\subsection{Implementation Details}
\noindent \textbf{Baseline}: We apply our method on PMM \cite{li2021pseudo}. It improves the generation and utilization of pseudo-masks based on SEAM \cite{Wang_2020_CVPR}. The backbone is ResNet-38 \cite{wu2019wider}, besides PMM uses Res2Net-101 \cite{gao2019res2net} and ScaleNet-101 \cite{li2019data} and achieve state-of-the-art results on both VOC and COCO. In this paper, we verify our methods on these three backbones and ResNet-101\cite{he2016deep}. Our settings are same to PMM. The different is that, we add our URN in this segmentation codebase. Specifically, the codebase is MMSegmentation \cite{mmseg2020}, and PMM uses PSPnet \cite{zhao2017pyramid} to get same results reported in the paper of SEAM. For VOC, the batch size is 16 on 8 GPUs at learning rate 0.005 for 20000 iterations in ploy policy. And training COCO requres 32 GPUSs at batch size 64 and learning rate 0.02. The iteration number is 40000. For the augmentation, the resized image is limited between 512 to 2048 with crop size 512$\times$512 for training. Besides, random flip and distortion are applied as transformation. The crop size of test phase is same as train phase, with dense-CRF as post-processing. For the classification part, use the original code, because we only focus on the segmentation phase. Note that, all the models are pretrained from imagenet.

\noindent \textbf{URN}: The weight mask is saved offline in PNG format to save storage space. For easily implementation, we concatenate pseudo-mask and weight mask into one image to aviod multiple ground-truths and multiple preprocessing. We split the weight mask and restore its range to 1 from 0 during rewight.
The scale factors $\boldsymbol{S}$ in Eq.(\ref{eq:scaling}) are set to \{0.15,0.2,0.25,4,5,6\} in VOC and \{0.4,0.5,0.6,2,3,4\} in COCO. The specific numbers are set by visualization without strict requirements. The minimum loss value $t$ which divide uncertain and uncertain pixels is determined by experiment at 0.05.

\noindent \textbf{Pseudo-mask Distillation}: As the segmentation model $\mathcal{F}_{S}$ is not the final deploy model, which returns prediction feature map and cyclic mask. Thus, we propose a practical and effective distillation scheme based on the cyclic pseudo-mask. If we have a teacher segmentation model $\mathcal{F}_{S}$ whose prediction $\boldsymbol{M}_{c}$ is in high quality, we train student models via this pseudo-mask for better guidance. In this paper, we set Res2Net-101 as the teacher model in VOC, and ScaleNet-101 is the teacher model in COCO. We also analyze the gains of Pseudo-mask Distillation in Tab.\ref{tab:distill}.

\subsection{Comparison with State-of-the-Art}
We conduct experiments on PASCAL VOC 2012 and MS COCO 2014 as Tab.\ref{tab:results}. The results in the top part are all trained from image-level annotations. And the works in the middle part introduce extra models or datasets. Except \cite{Khoreva_2017_CVPR} we don't list the methods supervised by bounding box, as we focus on image-level supervision methods. We organize the results by datasets and compare our method to the previous state-of-the-arts as follows.

\noindent \textbf{PASCAL VOC 2012}: We list the results on both validation set and test set on VOC. The most used backbone is ResNet-101, so we divide these results from others by color. For ResNet-101 our URN surpass the works in 2020 more than 3\% at 69.5\% on validation set. Another prevalent backbone is ResNet-38 which is wider than ResNet-101. Our result on this backbone is 70.6\% on test set. It is higher than our baseline PMM by 1.6\% and is about 4\% higher than CONTA which iterates 3 times on both classification and segmentation. Also, it is 4.9\% higher than SEAM. For the two stronger backbones ScaleNet-101 and Res2Net-101, we achieve 70.1\% and 71.2\% respectively.

\noindent \textbf{MS COCO 2014}: Compare to VOC, COCO is more challenging. There are 4 times categories and the training images are 8 times more, also the object size is smaller than VOC. Due to these difficulties, a few of works report the results on it, and there are results from validation set only before. For ResNet-38, SEAM's result is 31.7\% and PMM is at 36.7\%, while our URN achieve SOTA at 40.5\%. As well as other backbones, our results are higher than 40.5\%, among them the best mIoU is 41.5\%.

\noindent \textbf{Comparison with Methods with Extra Information}: In the middle part of Tab.\ref{tab:results}, we list some latest methods based on saliency, object proposals or extra datasets. We can see that these works are higher than the methods without extra information in average. But our results still beyond them at every dataset and backbone. Especially, URN is 7.1\% higher than SGAN on ResNet-101 on COCO, which demonstrates the strong effectiveness of our method.

\subsection{Ablation Studies}
\textbf{Effectiveness of URN}:
To verify the effectiveness of our method, we set experiments based on our baseline PMM which is the current state-of-the-art. We want to show that, our method works well even on a high baseline. The evaluated dataset is PASCAL VOC 2012 on the validation set with the metric of mIoU. We compare the results between baseline and ours on four backbones in Tab.\ref{tab:baseline}. It shows that our method achieves significant improvements on the all backbones. We push the previous state-of-the-art to 71.2\% from 70.0\%, and the gains on ResNet-101, ResNet-38 are 1.4\% and 0.9\% respectively. Among them, ScaleNet-101 raise most with a growth of 2.9\%. These results suggest that our method is effective and it works well on varied backbones.

\begin{table}[ht]
\begin{centering}
\setlength\tabcolsep{15pt}
\begin{tabular}{ccc}
\hline 
Backbone  & Baseline  & Ours\tabularnewline
\hline 
\hline 
ResNet-38  & 68.5  & 69.4\tabularnewline
ResNet-101  & 68.1  & 69.5\tabularnewline
ScaleNet-101  & 67.2  & 70.1\tabularnewline
Res2Net-101  & 70.0  & 71.2\tabularnewline
\hline 
\end{tabular}
\par\end{centering}
\caption{\label{tab:baseline}Comparison with baselines on PASCAL VOC 2O12 validation set.}
\end{table}

\noindent \textbf{Weight Threshould}: There is an important hyper paramter in URN which controls the intensity of reweight. To select the optimal value of $t$ in Eq.(\ref{eq:end}), we search it via experiments as shown in Tab.\ref{tab:thresh}. The backbone is Res2Net on PASCAL VOC 2012 validation set. If $t$ is set to 1, the weights are same to original weights, and 0 means droping all the uncertain pixels. It suggets that the pixels under uncertain area are useful, thus it is not wise to drop all of them. 0.05 is the optimal value, it means the weights of certain pixels is 20 times than the uncertain pixels, and it works best.
\begin{table}[htb]
\begin{centering}
\setlength\tabcolsep{15pt}
\begin{tabular}{cc}
\hline
$t$  & mIoU\tabularnewline
\hline
\hline
1  & 70.0\tabularnewline
0.5  & 70.3\tabularnewline
0.1  & 70.8\tabularnewline
0.05  & \textbf{71.2}\tabularnewline
0  & 70.2\tabularnewline
\hline
\end{tabular}
\par\end{centering}
\caption{\label{tab:baseline}Selection of threshold $t$ in Eq.(\ref{eq:end}).}
\label{tab:thresh}
\end{table}

\noindent \textbf{Pseudo-mask Distillation}:
Distillation is a widely used technique. In this paper we apply the mechanism by a very simple but practical way. We select the pseudo-masks whose quality are best as teacher and train other student models via them. Since student models are the deployed models, and they are able to learn the knowledge from teacher masks without increasing inference time. We verify the gains of Pseudo-mask Distillation in Tab. \ref{tab:distill}. We can see that the ScaleNet-101 and Res2Net-101 both raise 1.2\% after applying URN without PD, and ScaleNet-101 raises 1.8\% from Pseudo-mask Distillation. Note that Res2Net is the teacher in VOC, thus there is no result with PD. 
\begin{table}[t]
\begin{centering}
\setlength\tabcolsep{10pt}
\begin{tabular}{cccc}
\hline 
Backbone  & Basline & without PD  & with PD\tabularnewline
\hline 
\hline 
ScaleNet-101  & 67.1  & 68.3 & 70.1\tabularnewline
Res2Net-101  & 70.0  & 71.2 & -\tabularnewline
\hline 
\end{tabular}
\par\end{centering}
\caption{\label{tab:distill}Results of Pseudo-mask Distillation (PD). Note: Res2Net is the teacher model without PD.}
\end{table}

\noindent \textbf{Probability vs. Uncertainty}: As the uncertainty estimation replies on the prediction of segmentation, and some reweight methods use probability as metric in tasks like classification. We try to use the probability as loss weight for comparison. Specifically, we detach the prediction feature map and apply softmax to make the probability as loss weight. As shown in Tab.\ref{tab:prob}, in the same settings, the mIoU on VOC validation is 70.5\%, while our URN is 71.2\%. So probability has positive correlation to noise, but URN works better. Because in WSSS there are many false pixels with high probability.
\begin{table}[t]
\begin{centering}
\begin{tabular}{ccc}
\hline 
Method  & Backbone & VOC val\tabularnewline
\hline 
\hline 
Probability & Res2Net-101 & 70.5\tabularnewline
URN & Res2Net-101  & 71.2\tabularnewline
\hline 
\end{tabular}
\par\end{centering}
\caption{\label{tab:prob}Comparison of two reweight schemes.}
\end{table}

\subsection{Visual Comparison}
\begin{figure}[h!]
\begin{centering}
\includegraphics[width=2cm,height=1.7cm]{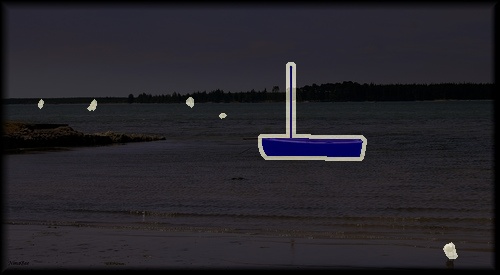} \includegraphics[width=2cm,height=1.7cm]{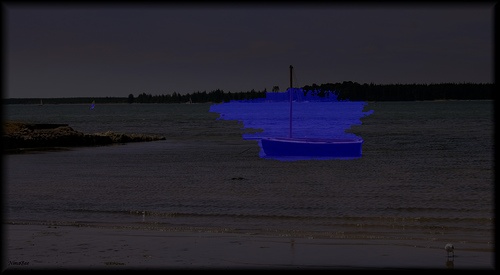} \includegraphics[width=2cm,height=1.7cm]{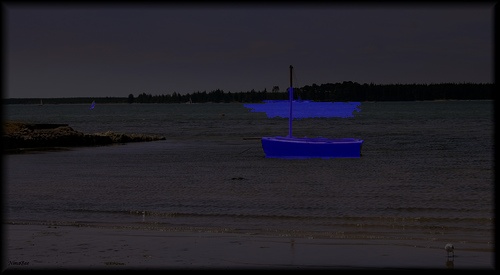} \includegraphics[width=2cm,height=1.7cm]{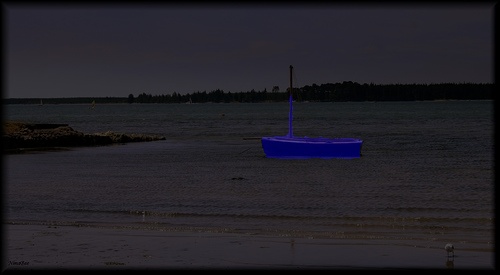}

\includegraphics[width=2cm,height=1.7cm]{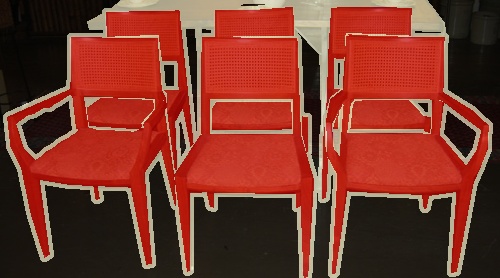} \includegraphics[width=2cm,height=1.7cm]{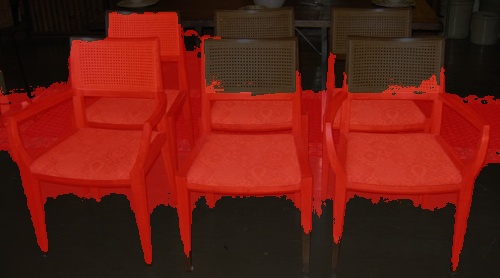} \includegraphics[width=2cm,height=1.7cm]{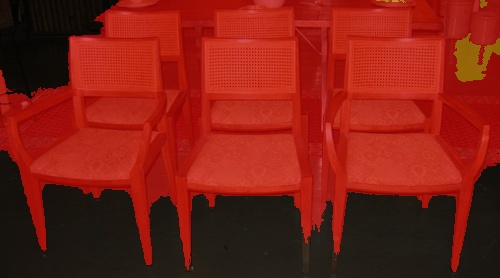} \includegraphics[width=2cm,height=1.7cm]{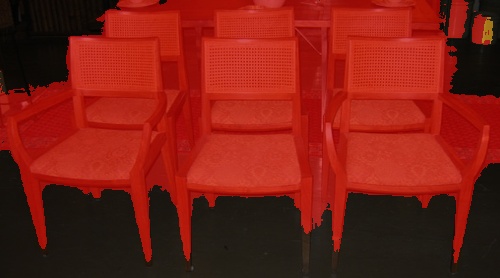}

\includegraphics[width=2cm,height=1.7cm]{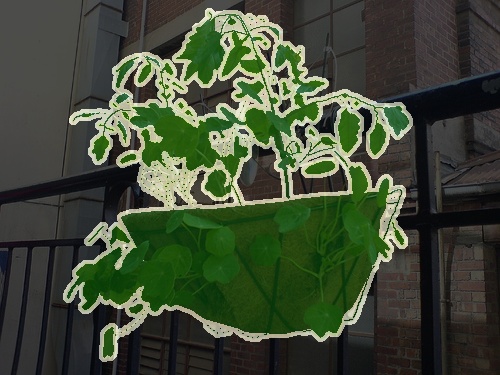} \includegraphics[width=2cm,height=1.7cm]{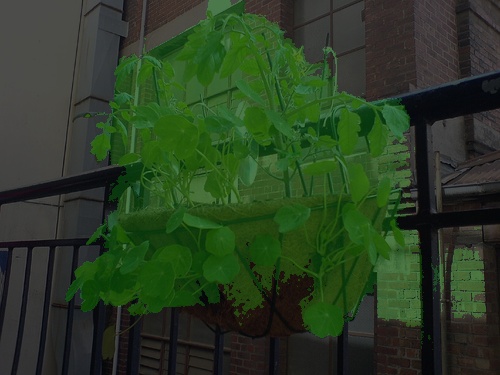} \includegraphics[width=2cm,height=1.7cm]{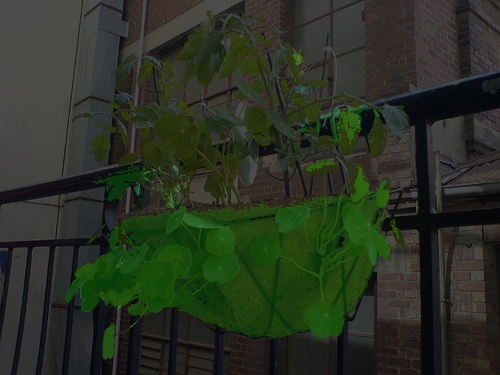} \includegraphics[width=2cm,height=1.7cm]{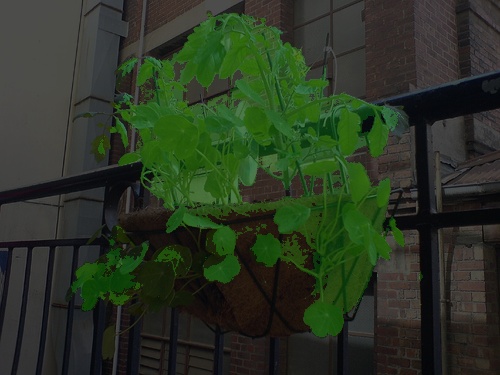}
  
\includegraphics[width=2cm,height=1.7cm]{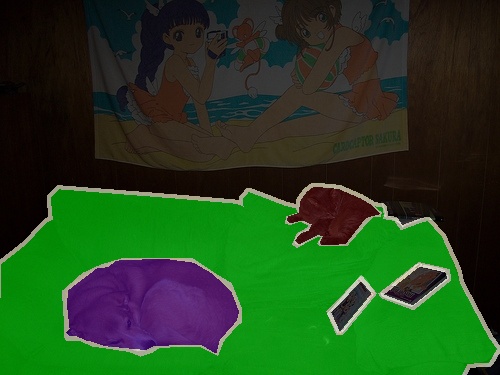} \includegraphics[width=2cm,height=1.7cm]{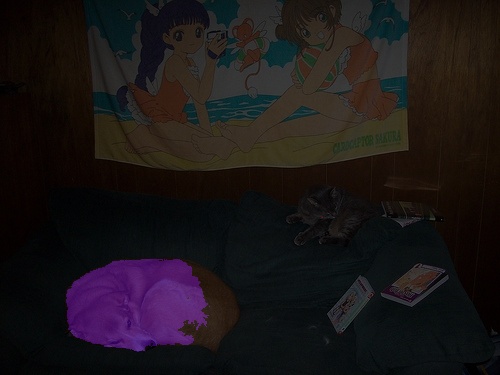} \includegraphics[width=2cm,height=1.7cm]{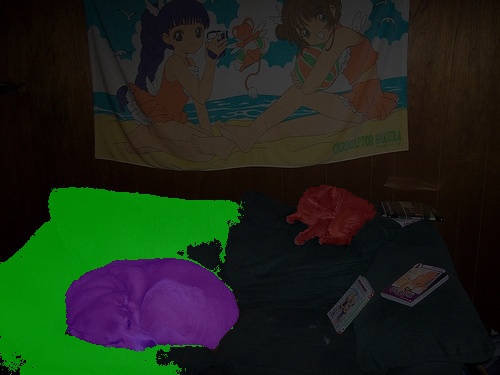} \includegraphics[width=2cm,height=1.7cm]{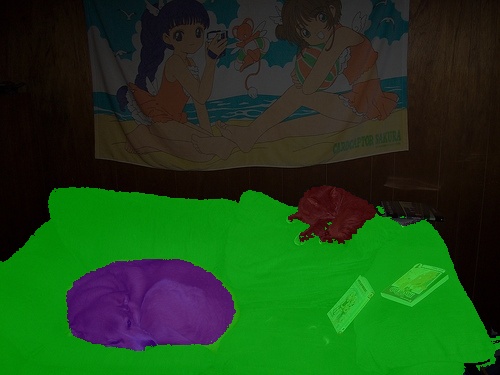}
\hspace*{0.25cm}GT\hspace*{1.2cm}SEAM\hspace*{1.2cm}PMM\hspace*{1.2cm}URN
\par\end{centering}
\begin{centering}
\caption{Visual comparison on PASCAL VOC 2012 validation set.}
\label{fig:voc_vis} 
\par\end{centering}
\centering{} 
\end{figure}

\begin{figure}[h!]
\begin{centering}
\includegraphics[width=2cm,height=1.7cm]{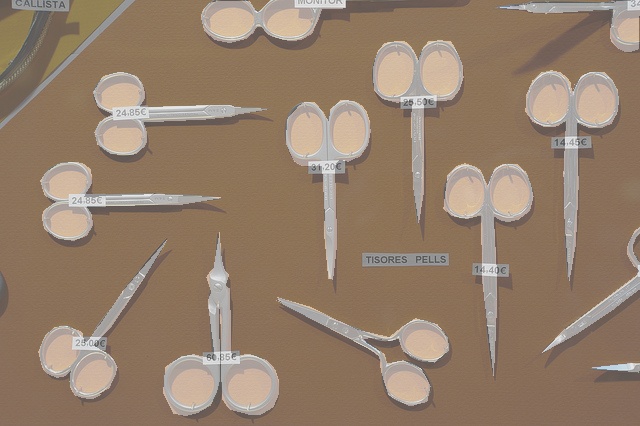} \includegraphics[width=2cm,height=1.7cm]{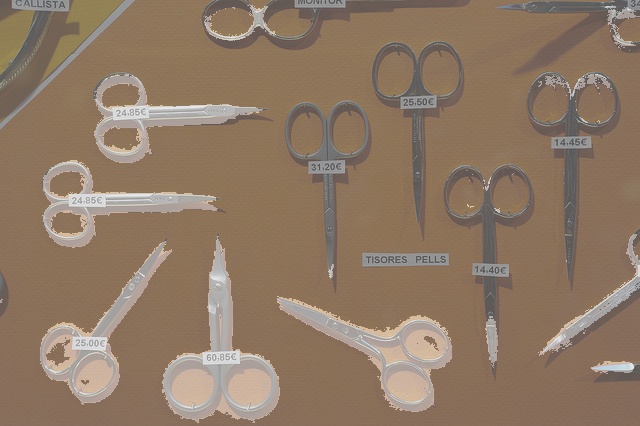} \includegraphics[width=2cm,height=1.7cm]{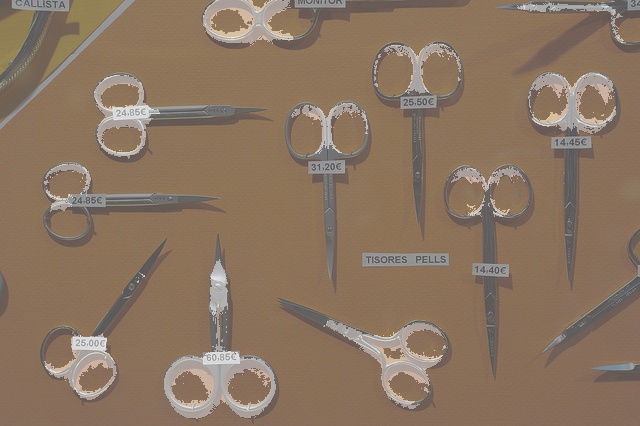}
\includegraphics[width=2cm,height=1.7cm]{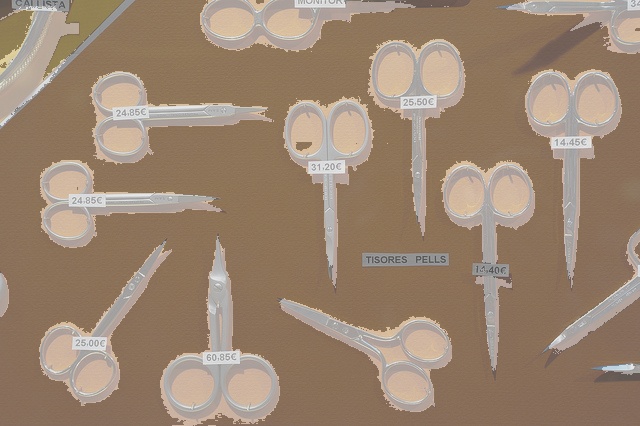}

\includegraphics[width=2cm,height=1.7cm]{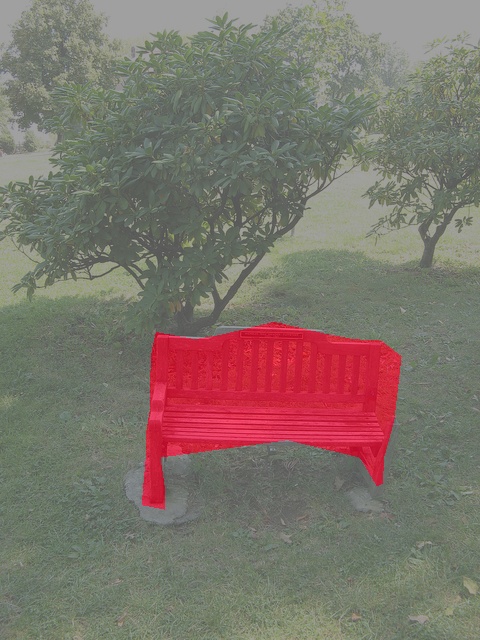} \includegraphics[width=2cm,height=1.7cm]{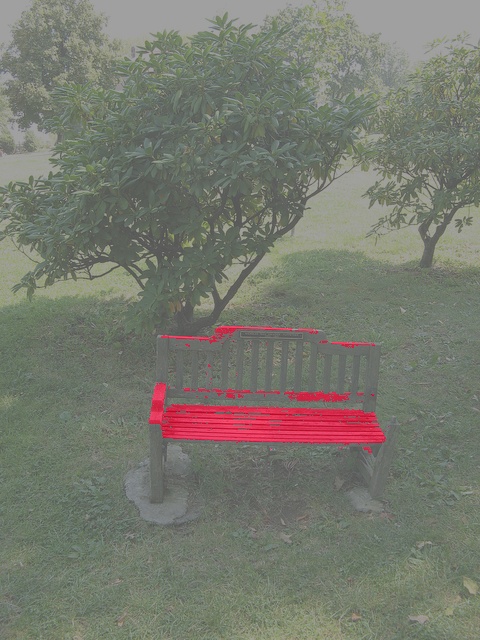} \includegraphics[width=2cm,height=1.7cm]{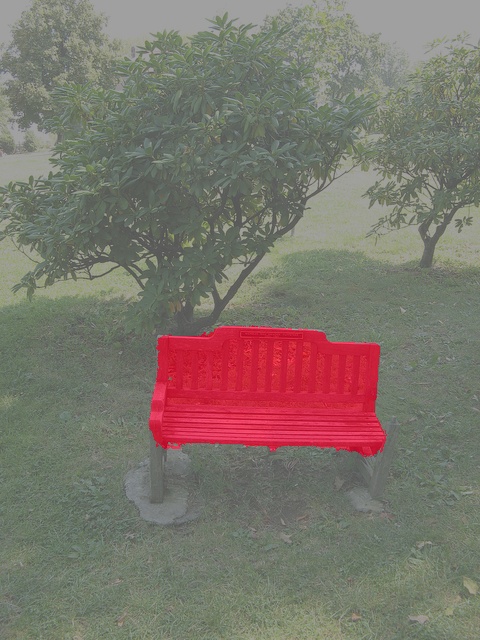}
\includegraphics[width=2cm,height=1.7cm]{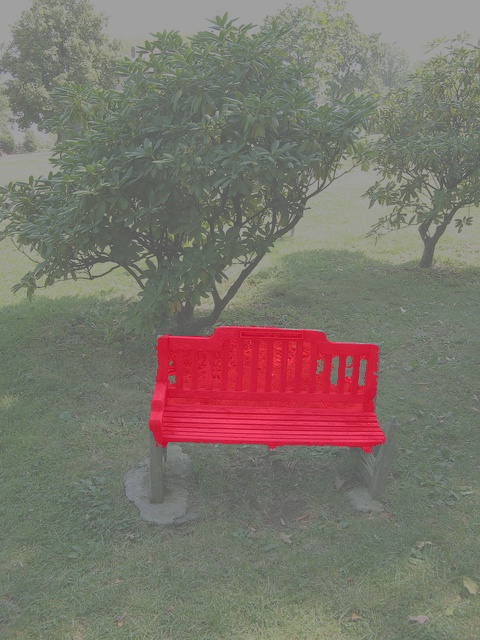}

\includegraphics[width=2cm,height=1.7cm]{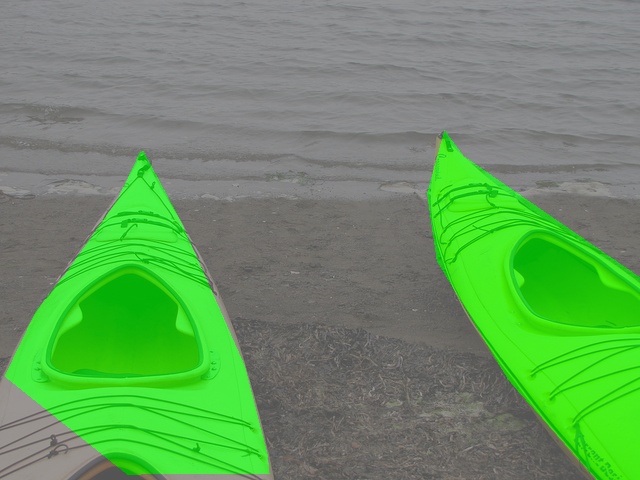} \includegraphics[width=2cm,height=1.7cm]{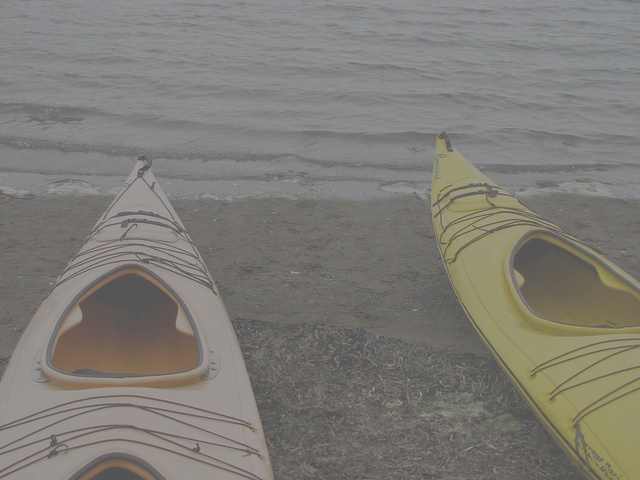} \includegraphics[width=2cm,height=1.7cm]{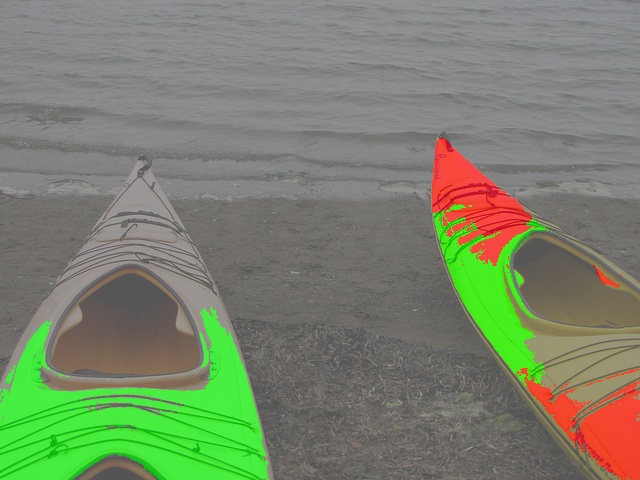}
\includegraphics[width=2cm,height=1.7cm]{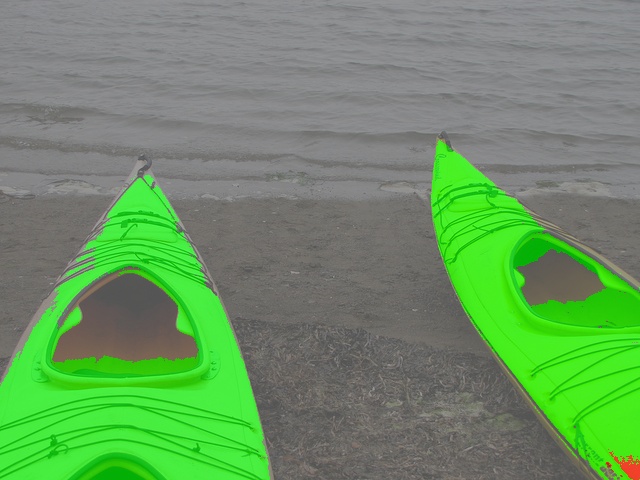}

\includegraphics[width=2cm,height=1.7cm]{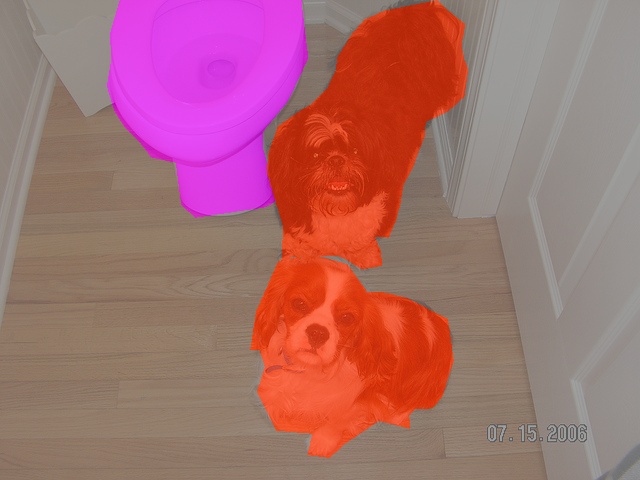} \includegraphics[width=2cm,height=1.7cm]{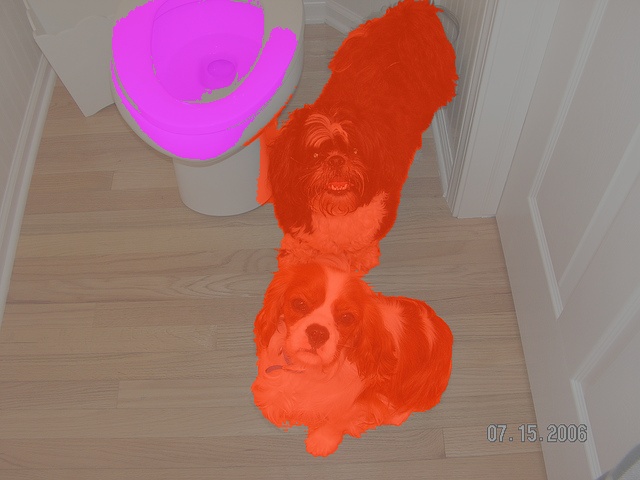} \includegraphics[width=2cm,height=1.7cm]{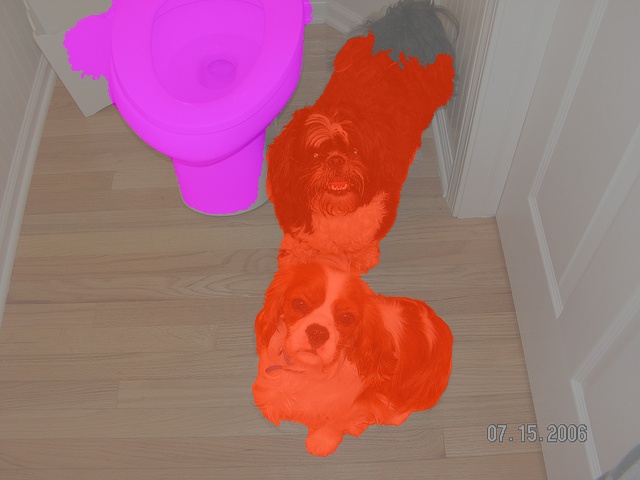}
\includegraphics[width=2cm,height=1.7cm]{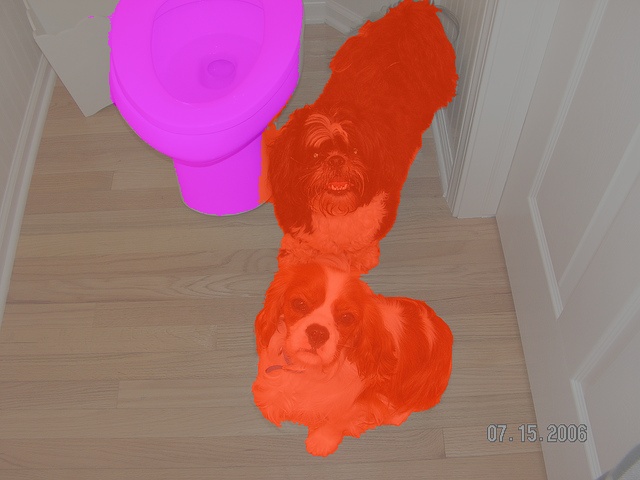}

\includegraphics[width=2cm,height=1.7cm]{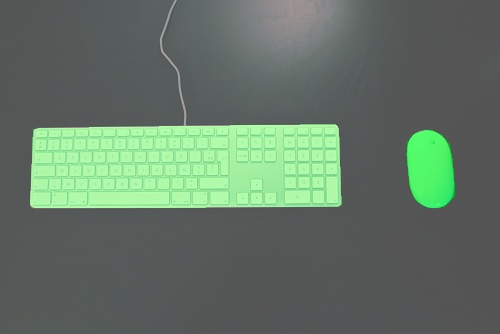} \includegraphics[width=2cm,height=1.7cm]{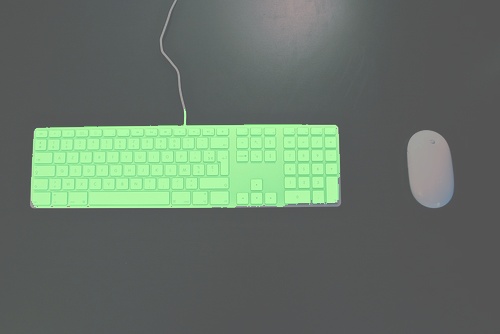} \includegraphics[width=2cm,height=1.7cm]{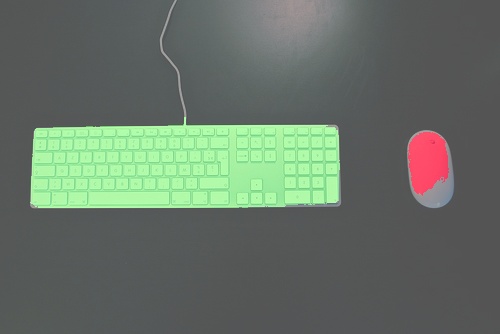}
\includegraphics[width=2cm,height=1.7cm]{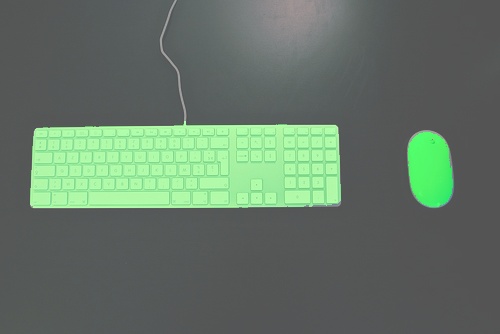}
\hspace*{0.25cm}GT\hspace*{1.2cm}SEAM\hspace*{1.2cm}PMM\hspace*{1.2cm}URN
\par\end{centering}
\begin{centering}
\caption{Visual comparison on MS COCO 2014 validation set.}
\label{fig:coco_vis} 
\par\end{centering}
\centering{} 
\vspace{-4mm}
\end{figure}

To verifies the effectiveness of our methods, we compare our results with SEAM and PMM on VOC12 validation set in Fig.\ref{fig:voc_vis}, and the visualizations of COCO are shown in Fig.\ref{fig:coco_vis}. In these two figures, we can see that our method performs better than these two baselines.

\section{Conclusion}

In summary, the noise is closely connected with the response scale, and we estimate the uncertainty by response scaling to simulate the various response scale. Then we mitigate the noise in segmentation optimization with the uncertainty from scaling. Besides, we later propose Pseudo-mask Distillation for the weaker backbones in implementation. Experimentally, we verify the improvements in obligation studies and compare our results to the previous state-of-the-art methods, which demonstrates the effectiveness of our method.

\section{Acknowledgement}
This work was supported by a research grant from Shenzhen Municipal Central Government Guides Local Science and Technology Development Special Funded Projects (2021Szvup139) and a research grant from HKUST Bridge Gap Fund (BGF.027.2021).


\clearpage 

{\small
\bibliography{aaai22}
}
\end{document}